\documentclass{article}

\usepackage[dblblindworkshop,final]{neurips_2026}

\usepackage[utf8]{inputenc}
\usepackage[T1]{fontenc}
\usepackage{hyperref}
\usepackage{url}
\usepackage{booktabs}
\usepackage{amsfonts}
\usepackage{amsmath}
\usepackage{nicefrac}
\usepackage{microtype}
\usepackage{xcolor}
\usepackage{graphicx}
\usepackage{caption}
\usepackage{multirow}
\usepackage[most]{tcolorbox}
\usepackage{tabularx}

\newcommand{\dm}{\Delta_M}

\newtcolorbox{finding}[1]{%
  enhanced,
  breakable,
  colback=blue!4!white,
  colframe=black,
  boxrule=0.8pt,
  arc=4pt,
  outer arc=4pt,
  left=6pt,
  right=6pt,
  top=5pt,
  bottom=5pt,
  before skip=6pt,
  after skip=6pt,
  attach boxed title to top left={
    xshift=6pt,
    yshift=-\tcboxedtitleheight/2
  },
  boxed title style={
    colback=black,
    colframe=black,
    boxrule=0pt,
    arc=2pt,
    outer arc=2pt,
    left=6pt,
    right=6pt,
    top=2pt,
    bottom=2pt,
  },
  fonttitle=\bfseries\footnotesize\color{white},
  fontupper=\small,
  title={#1}
}
\workshoptitle{Linguistic Principles for Foundation Models (LP4FM)}
\title{{Causal Syntax Circuits in CodeLMs: Budget and Probe Decide What Looks Localised}}

\author{%
  Harsh Kumar\thanks{Equal contribution.} \\
  Columbia University \\
  \texttt{hk3510@columbia.edu}
  \And
  Tanmay Shailesh Joshi\textsuperscript{*} \\
  BITS Pilani, K. K. Birla Goa Campus \\
  \texttt{tanmay7925@gmail.com}
  \And
  Rahul Maity\textsuperscript{*} \\
  Independent researcher \\
  \texttt{rahulmaity9977@gmail.com}
  \And
  Anusa Saha\textsuperscript{*} \\
  National Institute of Technology Agartala \\
  \texttt{anusasaha2406@gmail.com} \\
  \texttt{anusa\_23ucs042@ug.nita.ac.in}
}

\begin{document}

\maketitle

\begin{abstract}
Code models write Python when the prompt does not say which language to use. We
ask a mechanistic question about that habit: is it produced by a small, findable
group of components inside the model? We run one pipeline---mean ablation,
activation patching, greedy circuit extraction, a random-control significance
test, and activation steering---over 55 model--language settings spanning ten
open code models (five base and five instruction-tuned), three target languages
(Rust, C\#, PHP), two prompt suites, and two component budgets. We report five findings. First, the preference is
large, grows with model size, and survives instruction tuning: it is larger on
every instruction-tuned checkpoint than on its base version, by 12\% on average.
Second, its measured size is mostly a property of the probe; replacing our
aligned prompts with XLCoST parallel functions takes the same measurement on the
same models from about $1.6$ to about $0.01$. Third, whether the preference looks
localised depends on how many components the search may keep: with a budget of
12, 1 of 15 settings yields a circuit that beats size-matched random controls
after multiple-comparison correction, and with a budget of 40, 6 of 15 do.
Fourth, interventions that a smaller-budget study reports as uniformly failing do
sometimes succeed: patching makes a Python prompt prefer target syntax in 2 of 45
settings, and steering changes the most likely next token in 9 of 45. Fifth,
every circuit whose effect is significant but points \emph{away} from the target
language comes from a setting where circuit extraction failed, which makes such
results a diagnostic rather than a phenomenon. The practical lesson is that a
claim about whether a behaviour is localised can turn on analysis choices that
papers rarely report. We release the full grid so those choices can be varied.
\end{abstract}

\section{Introduction}
\label{sec:intro}

Ask a code model to ``sort a list'' and it will usually hand you Python. Ask it
in a Rust repository, with a Rust file open, and it will often still reach for
Python idioms unless the prompt spells out the language. Benchmarks report the
same pattern at scale: base code models score higher on Python than on the other
languages they were trained on, and they fall back to Python when the target is
left open \citep{cassano2023multipl,zheng2023codegeex}. We call this tendency
\emph{programming-language defaultness}. The two facts are separate, and we do
not claim one causes the other; the most likely common cause is that Python is
over-represented in code training corpora.

\paragraph{What behaviour would count as correct.} We are not arguing that a
Python default is a bug: when a task admits many languages, a model has to break
the tie somehow. The property we care about is narrower and testable---a user
should be able to \emph{move} the default---so our question is mechanistic, not
normative. Is the preference carried by a small group of components that an
inference-time edit could redirect, in the way that refusal
\citep{arditi2024refusal} and natural-language identity
\citep{tang2024langneurons,kojima2024multilingual} have been reported to be? A yes
would make cheap, weight-free language control possible for teams whose codebases
are not Python. A no tells practitioners to spend their effort elsewhere.

\paragraph{Why revisit this.} An earlier version of this study answered that
question in the negative across 15 model--language settings: the preference was
measurable everywhere, a specific circuit survived correction in one setting, and
neither patching nor steering ever changed the generated language. That conclusion
rested on two choices that looked like implementation details---keeping the top 12
components after ablation ranking, and measuring on a hand-built suite of aligned
function pairs---plus a scope restriction to base models.

\paragraph{Contributions.} We extend the grid from 15 settings to 55 by adding
(i) five instruction-tuned checkpoints, (ii) a component budget of 40, matching
the scale used by sparse-feature circuit work \citep{marks2025shift}, and (iii) a
second prompt suite drawn from XLCoST parallel functions
\citep{zhu2022xlcost}.\footnote{Code, configuration files, prompt suites, and the
result files for all 55 runs are available at
\url{https://github.com/Kitsunnneee/causal-syntax-circuits}.}
Table~\ref{tab:scorecard} is the whole result in one place, and each of the five
findings below is stated once, in a box, where its evidence is presented. Changing
those three things changes the conclusion.

\begin{table}[t]
  \caption{Every claim in the paper, under every probe: the number of settings
  supporting the claim out of the settings run. Only the first row is stable.
  The rest move when the component budget or the prompt suite changes, which is
  the paper's main point. Definitions are in \S\ref{sec:setup}.}
  \label{tab:scorecard}
  \centering
  \small
  \begin{tabularx}{\linewidth}{@{}Xcccc@{}}
\toprule
Claim & aligned, $k$=12 & aligned, $k$=40 & instruct, $k$=12 & XLCoST, $k$=12 \\
\midrule
$\operatorname{sign}\dm$ recovers the tagged language
& 15/15 & 15/15 & 15/15 & 0/10 \\

circuit beats random controls ($q_{\mathrm{BH}}<0.05$)
& 1/15 & 6/15 & 2/15 & 0/10 \\

circuit is significant but away-pointing
& 2/15 & 2/15 & 1/15 & 0/10 \\

patching lifts $\dm$ above zero
& 0/15 & 1/15 & 1/15 & 0/10 \\

steering makes the argmax target syntax
& 0/15 & 6/15 & 3/15 & 0/10 \\

\bottomrule
\end{tabularx}
\end{table}

\section{Localising behaviour, and what it has meant for language identity}

\paragraph{Localising behaviour by intervening.}
Our methods come from causal mediation analysis \citep{vig2020causal} and
activation patching \citep{meng2022rome}: a behaviour is traced by measuring how
a metric moves when internal activations are removed or transplanted between
runs. Circuit-level analyses in this tradition include the
indirect-object-identification circuit in GPT-2 \citep{wang2023ioi} and automated
circuit discovery \citep{conmy2023acdc,syed2024attribution}. We follow reported
best practice for patching metrics \citep{zhang2024bestpractices}, and our greedy
extraction approximates edge attribution patching \citep{syed2024attribution} at
the node level. Because such interventions can read structure into arbitrary
substructure, causal-abstraction work \citep{geiger2021causal,geiger2024das} and
interpretability surveys \citep{bereska2024mechanistic} stress the need for
controls, which is why every localisation claim here is gated on a comparison
against size-matched random component sets.

\paragraph{Directions and steering.}
A parallel line holds that concepts are encoded as directions rather than as
single neurons \citep{elhage2022superposition,park2024linear}, and that adding
such a direction during the forward pass steers behaviour
\citep{turner2023actadd,zou2023repe,li2023iti}. Single directions reportedly
mediate behaviours as specific as refusal \citep{arditi2024refusal}, and small
component sets can carry a task as a vector
\citep{todd2024functionvectors,hendel2023taskvectors}. Our steering estimator is
the standard difference-in-means construction; our contribution is to calibrate
it.

\paragraph{Language identity, natural and programming.} The closest precedent is
natural-language identity: a small set of ``language-specific neurons'',
concentrated in the first and last layers, can be toggled to change a multilingual
model's output language \citep{tang2024langneurons,kojima2024multilingual}. That
literature ties output-language preference to training-data frequency, and Python
defaultness is the code-side analogue. Programming languages differ in ways that
matter for whether the mechanism transfers: they share far more surface form with
each other than natural languages do, so identity is signalled by many correlated
cues and often unresolved until several tokens in. Whether a compact, steerable
mechanism exists here is therefore an open question rather than a corollary, and
to our knowledge this work and its earlier version are the first causal,
cross-model attempts to settle it. Our corpora come from MultiPL-E
\citep{cassano2023multipl}, XLCoST \citep{zhu2022xlcost}, and CodeXGLUE
\citep{lu2021codexglue}. Sparse-autoencoder decompositions
\citep{cunningham2024sae,bricken2023monosemanticity,templeton2024scaling} are an
alternative unit of analysis we return to in \S\ref{sec:discussion}.

\section{Measuring language preference}
\label{sec:setup}

\subsection{Prompts}
\label{sec:prompts}

Two prompt suites appear in this paper. Both consist of \emph{pairs}: a base
member written in Python and a target member written in Rust, C\#, or PHP, which
are identical in what the code does and differ in language.

\paragraph{Aligned prompts.} Our main suite is 64 structurally aligned function
pairs built from four templates (\textsc{add}, \textsc{square}, \textsc{greet},
\textsc{is\_even}). Each prompt is a one-shot context whose final line is a
language tag, so the model's next token is the keyword that opens a function
body. The base member is

\begin{quote}\ttfamily\small
\# language: python\\
def add(a, b):\\
\phantom{xx}return a + b\\
\# language: python\\
$\rightarrow$ \textrm{(next token scored here)}
\end{quote}

\noindent and the target member is identical up to the tag and body, e.g.\
\texttt{// language: rust} followed by \texttt{fn add(a: i32, b: i32) -> i32 \{
... \}}. The design does one job: it forces a language decision at a
known position, so the metric below is read where the model must commit. This
makes most other candidate tokens implausible there---after
\texttt{\# language: python} the plausible continuations are essentially
\texttt{def} and a newline---which is what makes the measurement clean and, as
Finding~2 shows, is also its main limitation.

\paragraph{XLCoST prompts.} To test whether the aligned suite manufactures its
own answer, we rerun the identical pipeline on 128 parallel function pairs drawn
from XLCoST \citep{zhu2022xlcost}, a cross-lingual code corpus of naturally
occurring programs aligned across languages. These prompts are real code, so the
scored position is wherever the aligned pair happens to end rather than a
constructed decision point.

\paragraph{Held-out behavioural prompts.} Generation and downstream checks use a
separate set of instruction prompts: two Python$\rightarrow$target translation
tasks, a target-language completion, and two controls (an English$\rightarrow$French
translation and a Python completion) that a syntax-specific intervention should
leave alone. Keeping these separate from the pairs stops the behavioural checks
contaminating the causal estimates.

\subsection{A continuous preference metric}
\label{sec:metric}

The language of a finished program is a discrete label, which gives an
intervention almost nothing to move. We instead read a continuous signal.
For a target language we fix a small set $V_{\mathrm{target}}$ of tokens that
open target-language syntax, and a set $V_{\mathrm{base}}$ that opens Python
syntax (Table~\ref{tab:vocab}). For an input $x$ we score

\begin{equation}
\label{eq:dm}
\dm(x) = \sum_{t \in V_{\mathrm{target}}} p(t \mid x) \;-\; \sum_{t \in V_{\mathrm{base}}} p(t \mid x) \;\in\; [-1, 1]
\end{equation}

\noindent at the scored position. Positive means the model is about to write the
target language; negative means Python. Surface strings are resolved to token ids
by encoding each under several natural prefixes (empty, leading space, newline,
newline plus indent) and keeping the first sub-token whose decoded form matches,
which recovers the id a model actually emits for, say, \texttt{fn} at the start
of a line. Ids appearing in both sets are dropped, so $\dm$ never credits a
language for syntax the two share. For a pair $(x_t, x_b)$ we summarise raw
preference by the \emph{separation} $S = \dm(x_t) - \dm(x_b)$.

\begin{table}[t]
  \caption{Syntax vocabularies defining $\dm$. Overlapping ids are removed at
  build time, so the two sets are disjoint. C\# and PHP need more strings because
  no single keyword is as diagnostic for them as \texttt{fn} is for Rust;
  \S\ref{sec:vocabsize} shows this does not drive the results.}
  \label{tab:vocab}
  \centering
  \small
  \begin{tabular}{lll}
    \toprule
    Language & Role & Tokens \\
    \midrule
    Rust   & target & \texttt{fn}, \texttt{mut}, \texttt{::} \\
    C\#    & target & \texttt{public}, \texttt{static}, \texttt{int}, \texttt{string}, \texttt{bool}, \texttt{;} \\
    PHP    & target & \texttt{function}, \texttt{\$}, \texttt{echo}, \texttt{array}, \texttt{foreach}, \texttt{;} \\
    Python & base   & \texttt{def}, \texttt{self}, \texttt{:} \\
    \bottomrule
  \end{tabular}
\end{table}

\subsection{Interventions, and the gate on every claim}

We study five model families across a $10\times$ size range---CodeLlama-7B
\citep{roziere2023codellama}, DeepSeek-Coder-1.3B/6.7B \citep{guo2024deepseek},
and Qwen2.5-Coder-7B/14B \citep{hui2024qwen}---in both base and instruction-tuned
form, against Rust, C\#, and PHP, through TransformerLens-style hooks
\citep{nanda2022transformerlens}. A component $c$ is an attention head or an MLP
block, addressed uniformly as a (kind, layer, head) triple. Let $a_c(x)$ be its
activation on input $x$. Four operators act on Eq.~\ref{eq:dm}.

\paragraph{Necessity.} Mean-ablate each component on the target prompt and rank
by the resulting drop, $\delta(c) = \dm(x_t) - \dm(x_t; a_c \!\leftarrow\! \bar a_c)$,
where $\bar a_c$ is the component's mean activation over the suite. We carry the
top $k$ components forward. Mean ablation, rather than zeroing, keeps the
statistics the rest of the network expects. \textbf{$k$ is the budget that
Finding~3 varies}: we run the whole grid at $k=12$ and again at $k=40$.

\paragraph{Sufficiency.} Transplant the target-run activations of a candidate set
$\mathcal{C}$ into the clean Python run and ask whether the sign changes:
$\dm^{\mathrm{patch}} = \dm(x_b;\, a_c \!\leftarrow\! a_c(x_t) \;\forall c \in \mathcal{C})$,
with success requiring $\dm(x_b) < 0 < \dm^{\mathrm{patch}}$.

\paragraph{Minimality.} Rank candidates by their individual patched shift and add
them greedily until the accumulated shift reaches $\tau = 0.90$ of the
full-set effect. This yields the extracted circuit and is a node-level
approximation of edge attribution patching \citep{syed2024attribution}.

\paragraph{Specificity, and the gate on every claim.} An extracted circuit is
only interesting if it beats a circuit of the same size chosen at random. We
compare the circuit's per-pair patched shifts against $N=30$ size-matched random
component sets with a Welch $t$-test. Because we run this test in every setting,
a raw threshold would over-report; we control the false-discovery rate with the
Benjamini--Hochberg procedure within each probe grid and require the corrected
$q_{\mathrm{BH}} < 0.05$. We say a circuit is \emph{specific} when it clears that
threshold \emph{and} its mean shift points toward the target language, and
\emph{away-pointing} when it clears the threshold but points the other way.

\paragraph{Steering.} For each circuit component we form the mean
target-minus-base direction and add it, scaled by $\alpha$, during the forward
pass on a Python prompt \citep{turner2023actadd}:
$\tilde a_c = a_c + \alpha\, d_c$ with $d_c = \mathbb{E}_x[a_c(x_t) - a_c(x_b)]$.
We sweep $\alpha \in \{0,1,2,4,8\}$ and record $\dm$, whether the most likely
next token is target syntax, and downstream generation behaviour.

\section{Results}
\label{sec:results}

\subsection{The preference survives scale and instruction tuning}
\label{sec:f1}

The sign of $\dm$ recovers the intended language in all 45 aligned-prompt
settings without exception: the target prompt always scores strongly positive
and the base prompt strongly negative. Separation grows with model size inside
each family, from a mean $S$ of $1.51$ at DeepSeek-Coder-1.3B to $1.61$ at
6.7B, and from $1.73$ at Qwen2.5-Coder-7B to $1.81$ at 14B. Two
sizes per family cannot establish a scaling law, but the direction is consistent.

\begin{figure}[t]
  \centering
  \includegraphics[width=\linewidth]{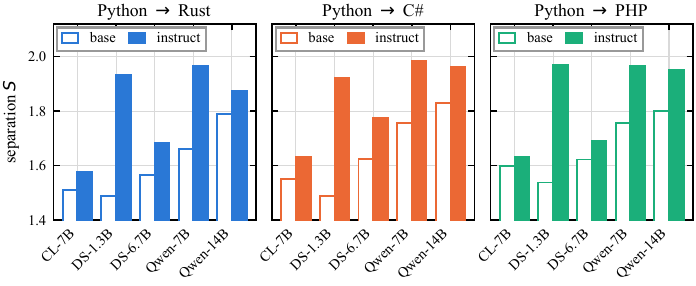}
  \caption{Separation $S$ for each base checkpoint (hollow bars) and its
  instruction-tuned sibling (filled), on the identical prompt suite. Instruction
  tuning increases $S$ in all 15 settings. Per-setting values are in
  Table~\ref{tab:preference}.}
  \label{fig:separation}
\end{figure}

Instruction tuning does not remove the preference. Figure~\ref{fig:separation}
pairs each base checkpoint with its instruction-tuned sibling on the identical
measurement. The instruction-tuned model has the \emph{larger} separation in all
15 settings, by 12\% on average and by as much as 30\% (DeepSeek-Coder-1.3B on
Rust, $1.49 \rightarrow 1.93$). The base Python score also becomes more negative
after tuning, meaning the tuned model is more confidently Python when the tag
says Python. This matters for motivation: whatever post-training does to these
models, it does not dissolve the phenomenon, so studying it on base models was
not a scope error.

Because $S$ is read at a language-\emph{tagged} position, a negative base score
could reflect tag-following instead of a real Python default. Re-measuring $\dm$
on ten bare instruction prompts with no tag (Appendix~\ref{app:untagged}) leaves
the default intact: it is Python-leaning in 12 of 15 settings, with the three
exceptions all PHP. The effect is one to two orders of magnitude smaller than
$S$, so the unconstrained prior is real but faint.

\begin{finding}{Finding 1}
Language preference is large, grows with model size, and is \emph{stronger}
after instruction tuning, not weaker.
\end{finding}
\vspace{-0.75em}

\subsection{Vocabulary size does not drive the metric}
\label{sec:vocabsize}

Because Eq.~\ref{eq:dm} sums probabilities, a larger $V_{\mathrm{target}}$ could
inflate $\dm$ mechanically, and the C\# and PHP sets are twice the size of the
Rust set after id resolution (12 ids vs.\ 6; the Python set is 6 in all cases).
Three observations bound this concern. First, the base term is identical across
the three target languages within a model, so only the target term can vary.
Second, doubling the vocabulary buys very little: averaged over the five base
models, mean $S$ is $1.603$ for Rust, $1.650$ for C\#, and $1.663$ for PHP---a
gap of $0.06$, or 3.7\% of $S$, against a $2\times$ difference in set size.
Third, across the 15 aligned base settings, set size and separation are not
associated (Spearman $\rho = +0.20$, $p = 0.48$). The scores are also near
saturation ($\dm(x_t) \in [0.76, 0.90]$), which caps how much extra mass more
tokens can collect. A fully size-matched vocabulary would still be the cleaner
design, and we flag it as unfinished work in \S\ref{sec:limitations}.

\subsection{The measured size belongs to the probe, not to the model}
\label{sec:f2}

Rerunning the identical pipeline on XLCoST parallel functions gives a different
picture of the same models. Separation collapses from $\approx 1.6$ to below $0.012$ in every one of the ten
settings---two orders of magnitude (Table~\ref{tab:preference}). Both terms of Eq.~\ref{eq:dm} are near zero: on
naturally occurring code, at the position where an aligned pair happens to end,
the model is not about to emit a language-defining keyword at all, so almost no
probability mass sits on either vocabulary.

We read this as a statement about the probe, not about the models. The aligned
suite does not measure defaultness in the wild; it constructs a position where a
language decision must be made and reads the model's commitment there. That is
what makes the causal interventions interpretable, but the number it produces is
not the size of the bias a developer meets. Consistent with this, no XLCoST
setting yields a specific circuit and no intervention moves anything: with the
metric already at zero there is nothing to move.

\begin{table}[t]
  \caption{Language preference under all four probes. $\dm(x_t)$ scores the
  target-language prompt and $S$ is the separation between the target and base
  prompts (\S\ref{sec:metric}). ``No tag'' is the mean $\dm$ over ten bare
  instruction prompts carrying no language tag, where negative means
  Python-leaning. XLCoST was run for C\# and PHP only.}
  \label{tab:preference}
  \centering
  \footnotesize
  \begin{tabular}{llrrrrr}
\toprule
\multirow{2}{*}{Model} & \multirow{2}{*}{Target} & \multicolumn{2}{c}{Aligned prompts, base} & \multicolumn{1}{c}{Instruct} & \multicolumn{1}{c}{XLCoST} & \multicolumn{1}{c}{No tag} \\
\cmidrule(lr){3-4}\cmidrule(lr){5-5}\cmidrule(lr){6-6}\cmidrule(lr){7-7}
 & & $\dm(x_t)$ & $S$ & $S$ & $S$ & mean (sd) \\
\midrule
CodeLlama-7B & Rust & +0.818 & 1.511 & 1.577 & --- & -0.010 (0.008) \\
 & C\# & +0.857 & 1.551 & 1.631 & +0.012 & -0.010 (0.008) \\
 & PHP & +0.904 & 1.597 & 1.632 & +0.001 & +0.022 (0.020) \\
\addlinespace[2pt]
DeepSeek-Coder-1.3B & Rust & +0.760 & 1.490 & 1.932 & --- & -0.040 (0.022) \\
 & C\# & +0.759 & 1.488 & 1.920 & -0.006 & -0.038 (0.023) \\
 & PHP & +0.809 & 1.538 & 1.968 & +0.003 & +0.013 (0.032) \\
\addlinespace[2pt]
DeepSeek-Coder-6.7B & Rust & +0.788 & 1.567 & 1.683 & --- & -0.070 (0.031) \\
 & C\# & +0.845 & 1.625 & 1.774 & +0.011 & -0.063 (0.033) \\
 & PHP & +0.844 & 1.623 & 1.692 & +0.007 & +0.076 (0.056) \\
\addlinespace[2pt]
Qwen2.5-Coder-7B & Rust & +0.750 & 1.660 & 1.965 & --- & -0.387 (0.171) \\
 & C\# & +0.848 & 1.758 & 1.984 & +0.004 & -0.381 (0.170) \\
 & PHP & +0.848 & 1.758 & 1.965 & +0.002 & -0.360 (0.171) \\
\addlinespace[2pt]
Qwen2.5-Coder-14B & Rust & +0.855 & 1.789 & 1.875 & --- & -0.250 (0.077) \\
 & C\# & +0.895 & 1.828 & 1.961 & +0.009 & -0.245 (0.077) \\
 & PHP & +0.867 & 1.801 & 1.949 & +0.000 & -0.215 (0.075) \\
\bottomrule
\end{tabular}

\end{table}

\begin{finding}{Finding 2}
The measured size of the preference belongs to the probe: the same metric on the
same models reads $\approx 1.6$ on aligned prompts and $\approx 0.01$ on XLCoST.
\end{finding}

\subsection{Localisation depends on the component budget}
\label{sec:f3}

Keeping the top 12 components after ablation ranking was never justified by
anything except tractability, yet it carries the earlier version's central
negative claim. Table~\ref{tab:specificity} gives the specificity test at both
budgets. At $k=12$, exactly one setting (CodeLlama-7B on Rust) beats size-matched
random controls after correction. At $k=40$---the range used by sparse-feature
circuit work \citep{marks2025shift}---six do, and two of them are large:
DeepSeek-Coder-1.3B/C\# shifts $\dm$ by $+0.61$ and Qwen2.5-Coder-7B/PHP by
$+1.25$, against random-control means below $0.03$. On the instruction-tuned grid
at $k=12$, two settings are specific, one of them ($+1.27$) larger than anything
on the base models.

Only CodeLlama-7B/Rust is specific at both budgets, so the identity of the
surviving settings is not stable either. The circuits share no structure across
models: they span the full depth range (\texttt{L9MLP} in CodeLlama-7B,
\texttt{L12MLP} in Qwen2.5-Coder-14B, \texttt{L22--L24MLP} in Qwen2.5-Coder-7B,
\texttt{L29MLP} in DeepSeek-Coder-6.7B), and they are MLP-mediated: attention
heads appear in the ablation rankings but rarely survive extraction. Circuit size
moves with the budget too---at $k=40$, four settings return the full candidate set
and four others return a single component (Appendix~\ref{app:circuits}).

The honest summary is not ``the preference is localisable'' or ``it is not''. It
is that a study reporting either answer is reporting a property of its search
budget as much as of the model, and that budgets in this literature are chosen by
convention.

\begin{table}[t]
  \caption{Specificity across component budgets. ``shift'' is the circuit's mean
  patched shift in $\dm$; $q_{\mathrm{BH}}$ corrects the Welch test against 30
  size-matched random sets within each column pair. $^{\ast}$ specific;
  $^{\dagger}$ significant but pointing away from the target language.}
  \label{tab:specificity}
  \centering
  \footnotesize
  \setlength{\tabcolsep}{4.5pt}
  \begin{tabular}{llrrrrrr}
\toprule
\multirow{2}{*}{Model} & \multirow{2}{*}{Target} & \multicolumn{2}{c}{Aligned, $k$=12} & \multicolumn{2}{c}{Aligned, $k$=40} & \multicolumn{2}{c}{Instruct, $k$=12} \\
\cmidrule(lr){3-4}\cmidrule(lr){5-6}\cmidrule(lr){7-8}
 & & shift & $q_{\mathrm{BH}}$ & shift & $q_{\mathrm{BH}}$ & shift & $q_{\mathrm{BH}}$ \\
\midrule
CodeLlama-7B & Rust & +0.066$^{\ast}$ & 0.030 & +0.066$^{\ast}$ & 0.016 & -0.025 & 0.528 \\
 & C\# & -0.412$^{\dagger}$ & 0.030 & -0.429$^{\dagger}$ & 0.011 & -0.351$^{\dagger}$ & 0.012 \\
 & PHP & -0.082$^{\dagger}$ & 0.040 & +0.027 & 0.124 & -0.050 & 0.080 \\
\addlinespace[2pt]
DeepSeek-Coder-1.3B & Rust & +0.010 & 0.670 & -0.040 & 0.659 & +0.040$^{\ast}$ & 0.012 \\
 & C\# & -0.019 & 0.131 & +0.612$^{\ast}$ & 0.000 & +0.057 & 0.134 \\
 & PHP & +0.050 & 0.131 & +0.049 & 0.094 & +0.014 & 0.171 \\
\addlinespace[2pt]
DeepSeek-Coder-6.7B & Rust & +0.056 & 0.109 & -0.023$^{\dagger}$ & 0.045 & +0.035 & 0.380 \\
 & C\# & +0.005 & 0.300 & +0.023$^{\ast}$ & 0.031 & +0.031 & 0.442 \\
 & PHP & -0.006 & 0.388 & +0.049$^{\ast}$ & 0.031 & +0.015 & 0.674 \\
\addlinespace[2pt]
Qwen2.5-Coder-7B & Rust & -0.023 & 0.418 & +0.008 & 0.593 & +1.270$^{\ast}$ & 0.012 \\
 & C\# & +0.017 & 0.418 & +0.036 & 0.398 & +0.000 & 0.674 \\
 & PHP & -0.011 & 0.131 & +1.250$^{\ast}$ & 0.000 & -0.011 & 0.528 \\
\addlinespace[2pt]
Qwen2.5-Coder-14B & Rust & -0.053 & 0.200 & -0.046 & 0.296 & -0.040 & 0.380 \\
 & C\# & -0.030 & 0.131 & +0.018$^{\ast}$ & 0.031 & -0.026 & 0.419 \\
 & PHP & +0.023 & 0.314 & +0.066 & 0.058 & +0.005 & 0.467 \\
\midrule
\multicolumn{2}{l}{\emph{specific circuits}} & \multicolumn{2}{c}{1/15} & \multicolumn{2}{c}{6/15} & \multicolumn{2}{c}{2/15} \\
\bottomrule
\end{tabular}

\end{table}

\begin{finding}{Finding 3}
Whether the preference looks localised depends on the search budget: going from
12 components to 40 takes the count of specific circuits from 1 of 15 to 6 of 15.
\end{finding}
\vspace{-0.75em}
\subsection{Patching and steering work in a minority of settings}
\label{sec:f4}

\paragraph{Patching.} Transplanting the extracted components from the target run
into a Python prompt moves $\dm$ across zero in 2 of the 45 aligned and
instruction-tuned settings (Table~\ref{tab:intervention}, left). Both are
Qwen2.5-Coder-7B: the base model on PHP at $k=40$ ($-0.91 \rightarrow +0.50$,
circuit \texttt{L22--L24MLP} plus \texttt{L27H13}) and the instruction-tuned model
on Rust at $k=12$ ($-0.99 \rightarrow +0.50$, a 12-component set led by
\texttt{L24MLP}). Both circuits are also specific by the test of
\S\ref{sec:f3}, so the two results agree. Elsewhere the patched metric stays
negative; the next closest is DeepSeek-Coder-1.3B/C\# at $-0.14$ and every other
setting stays below $-0.6$, so these are not near misses.

\begin{figure}[t]
  \centering
  \includegraphics[width=\linewidth]{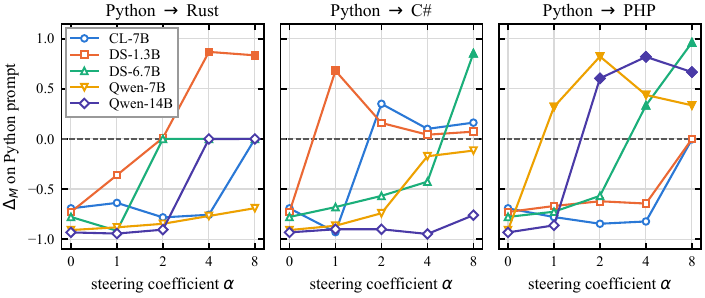}
  \caption{Steering sweeps at $k=40$ on the base models, one panel per target
  language. Filled markers are coefficients at which the most likely next token
  has become target syntax. The metric moves well before the argmax does, and in
  most settings the argmax never follows.}
  \label{fig:steering}
\end{figure}

\paragraph{Steering.} Adding the difference-in-means direction changes the most
likely next token to target syntax in 9 of 45 settings: 6 at $k=40$ on base models
and 3 on instruction-tuned models (Table~\ref{tab:intervention}, right;
Figure~\ref{fig:steering}). At $k=12$ on base models it never does, which is what
the earlier version reported. The coefficient matters. Three of the nine switch at
$\alpha=1$, two of them the settings where patching also worked; the remaining six
need $\alpha \geq 2$ and four need $\alpha \geq 4$. That upper range is where
behaviour degrades: generation becomes repetitive, and on Qwen2.5-Coder-14B/C\# a
coefficient of 4 or more injects 126--128 C\# markers into control prompts that
should contain none. A safe coefficient exists but must be searched for per
setting: for Qwen2.5-Coder-7B it is 8 on Rust, 4 on PHP, and 0 on C\#, so C\#
cannot be steered safely at any strength we test (Appendices~\ref{app:safe}, \ref{app:contam}).

\paragraph{Reading a metric flip correctly.} A curve crossing zero is tempting to
read as a switch, but $\dm$ is a difference of probability masses: it rises
either by moving mass onto target keywords or by flattening the distribution.
Two checks separate these. The argmax check---is the most likely token target
syntax?---is the one reported above. The contamination check asks whether the
same coefficient injects target markers into unrelated prompts, and it fails for
Qwen2.5-Coder-14B/C\# at $\alpha \geq 4$. A defensible claim needs a metric shift,
a target argmax, and clean controls together. Nine settings give the first two;
the coefficient window in which they give all three is narrow and
model-specific.

\begin{table}[t]
  \caption{Interventions on the Python prompt. Left: $\dm$ after patching the
  extracted circuit; the clean value before patching is $\dm(x_t)-S$ from
  Table~\ref{tab:preference}, and bold marks the two runs that cross zero.
  Right: the smallest steering coefficient at which the most likely next token
  becomes target syntax; ``---'' means none of $\{1,2,4,8\}$ does.}
  \label{tab:intervention}
  \centering
  \footnotesize
  \setlength{\tabcolsep}{4.5pt}
  \begin{tabular}{llrrrrrr}
\toprule
\multirow{2}{*}{Model} & \multirow{2}{*}{Target} & \multicolumn{3}{c}{Patched $\dm$ (clean $\rightarrow$ patched)} & \multicolumn{3}{c}{Smallest $\alpha$ with target argmax} \\
\cmidrule(lr){3-5}\cmidrule(lr){6-8}
 & & $k$=12 & $k$=40 & instr. & $k$=12 & $k$=40 & instr. \\
\midrule
CodeLlama-7B & Rust & -0.686 & -0.692 & -0.761 & --- & --- & --- \\
 & C\# & -0.979 & -0.987 & -0.991 & --- & --- & --- \\
 & PHP & -0.738 & -0.668 & -0.751 & --- & --- & 2 \\
\addlinespace[2pt]
DeepSeek-Coder-1.3B & Rust & -0.718 & -0.742 & -0.924 & --- & 4 & 4 \\
 & C\# & -0.752 & -0.136 & -0.939 & --- & 1 & --- \\
 & PHP & -0.695 & -0.637 & -0.970 & --- & --- & --- \\
\addlinespace[2pt]
DeepSeek-Coder-6.7B & Rust & -0.732 & -0.811 & -0.790 & --- & --- & --- \\
 & C\# & -0.773 & -0.745 & -0.797 & --- & 8 & --- \\
 & PHP & -0.793 & -0.739 & -0.854 & --- & 4 & --- \\
\addlinespace[2pt]
Qwen2.5-Coder-7B & Rust & -0.914 & -0.898 & \textbf{+0.504} & --- & --- & 1 \\
 & C\# & -0.859 & -0.863 & -0.992 & --- & --- & --- \\
 & PHP & -0.906 & \textbf{+0.500} & -0.992 & --- & 1 & --- \\
\addlinespace[2pt]
Qwen2.5-Coder-14B & Rust & -0.949 & -0.945 & -0.980 & --- & --- & --- \\
 & C\# & -0.949 & -0.914 & -0.973 & --- & --- & --- \\
 & PHP & -0.891 & -0.840 & -0.965 & --- & 2 & --- \\
\bottomrule
\end{tabular}

\end{table}

\begin{finding}{Finding 4}
Patching and steering are not uniformly inert---they work in 2 of 45 and 9 of 45
settings---but mostly at coefficients large enough to damage the output.
\end{finding}
\vspace{-0.75em}
\subsection{No single component is behaviourally load-bearing}

Starting from the CodeLlama-7B/Rust set at $\alpha=1$, we remove one component at
a time and re-score the explicit Rust and control instruction suite. Every
removal---including dropping the specific circuit itself---leaves marker accuracy
and F1 at $1.000$, and mean next-token $\dm$ moves by less than $0.02$
(Appendix~\ref{app:loo}). Explicit prompts carry so many correlated language cues
that no single steering component changes what the model does. Two CodeXGLUE tasks
\citep{lu2021codexglue} stay pinned at chance under steering, so the direction
encodes surface syntax, not task competence (Appendix~\ref{app:xglue}).

\subsection{Circuits that point away are extraction failures}
\label{sec:wrongdir}

Five results across three settings clear the significance threshold with the
shift pointing the wrong way; the largest, CodeLlama-7B on C\# at $-0.43$,
reproduces at both budgets and on the instruction-tuned checkpoint. Transplanting
C\# activations into a Python prompt makes the model \emph{more} confidently
Python. Two observations explain it, and neither is about C\#.

First, transplanting a late-layer activation is damaging in every language.
Figure~\ref{fig:depth} sums each component's individual patched shift by depth
band in CodeLlama-7B. Early components contribute close to nothing ($+0.06$,
$+0.01$, $+0.08$ for Rust, C\# and PHP); middle and late components contribute
large negative shifts, the late band with only four or five components. A
late-layer activation is computed for a specific context---part-way through a
target-language function body---so transplanting it into a prompt sitting just
after a Python language tag inserts a state the network never produces, and the
model falls back to its most confident generic continuation, which after
\texttt{\# language: python} is Python syntax. Summed over all 40 components the
damage is nearly identical across the three languages ($-0.46$, $-0.43$,
$-0.46$), so it cannot be what separates them.

Second, what separates them is whether extraction keeps those components. For
CodeLlama-7B, greedy extraction returns the single component \texttt{L9MLP} for
Rust and for PHP, but for C\# it finds no compact subset and returns all 40
candidates, including every damaging late MLP (Table~\ref{tab:circuits}). The
pattern holds across the grid: all five away-pointing results come from settings
where extraction returned the full candidate set, and at $k=40$ full-set circuits
have a median shift of $-0.04$ against $+0.05$ for compact ones. A significant
negative result is therefore a warning about the search, not evidence of an
anti-circuit, and such searches should report extracted size alongside
significance.

\begin{finding}{Finding 5}
Every significant \emph{away}-pointing circuit comes from a setting where
extraction failed to find a compact subset. It diagnoses the search, not the model.
\end{finding}
\vspace{-0.75em}
\section{Discussion}
\label{sec:discussion}

\paragraph{For practitioners.} The preference is real and cheap to monitor---one
forward pass---but it is not a reliable dial. Steering changes the next token in a
fifth of settings, needs a coefficient searched per setting, and trades syntax for
fluency past that point. Prompting and fine-tuning remain the dependable route.

\paragraph{For interpretability research.} This is the result we think travels.
Two knobs normally reported as setup---how many components the search keeps, and
which prompts the metric is read on---moved our headline conclusions from ``not
localisable, never steerable'' to ``localisable in 6 of 15 settings, steerable in
9 of 45'', and from ``a large, robust bias'' to ``a bias two orders of magnitude
smaller''. Neither knob is exotic. Circuit papers should report a budget sweep the
way ablation papers report a seed sweep, and state a localisation claim as a curve
over budgets rather than a single number. Where a handful of neurons can toggle a
multilingual model's output language
\citep{tang2024langneurons,kojima2024multilingual}, we find no comparably compact
structure for code: identity there is signalled by many correlated cues, so no
component is a bottleneck. Sparse-feature \citep{cunningham2024sae,marks2025shift}
or edge-level \citep{syed2024attribution} analyses are the natural next step.

\section{Limitations}
\label{sec:limitations}

\textbf{Metric.} $\dm$ uses a small hand-chosen vocabulary read at one position;
\S\ref{sec:vocabsize} bounds the effect of set size without removing it, and the
C\# and PHP sets share markers with other C-family languages. \textbf{Method.}
Minimality is node-level and approximates edge attribution patching, and steering
adds one difference-in-means direction, so Finding~4 bounds these methods rather
than all possible controls. \textbf{Scope.} We test two budgets, one base
language, three targets, and two sizes per family---enough to show that budget
matters, not enough for a scaling law.

\section{Conclusion}

Code models' preference for Python is large, grows with scale, and survives
instruction tuning, but the mechanistic story about it is not stable under choices
that look like setup. Raising the budget from 12 components to 40 takes the number
of settings with a statistically specific circuit from 1 to 6; swapping the prompt
suite takes the measured preference from $1.6$ to $0.01$; and every away-pointing
circuit marks a failed search. We release the grid and the pipeline so these
sweeps can be continued rather than assumed.

\bibliographystyle{plainnat}
\bibliography{refs}

\appendix
\section*{Appendix}

\section{Configuration and implementation}
\label{app:config}

All runs share the settings in Table~\ref{tab:config}. CodeLlama uses a
TransformerLens \texttt{HookedTransformer} backend; DeepSeek-Coder and
Qwen2.5-Coder, which are not in that registry, use the TransformerLens bridge
backend, which wraps a Hugging Face model and re-exposes the same
\texttt{blocks.\{layer\}.attn.hook\_z} and \texttt{blocks.\{layer\}.hook\_mlp\_out}
hook points, so one implementation of every intervention runs unchanged across
backends. Attention-head interventions act on the per-head \texttt{hook\_z}
tensor and MLP interventions on \texttt{hook\_mlp\_out}. The mean-ablation cache
is built once per run by averaging each hook's activation over the prompt suite
with the position axis reduced, so a single stored vector broadcasts over any
sequence length. Qwen2.5-Coder produced NaN logits in fp16 on the bridge path, so
all Qwen runs use bf16 and the others use fp16. Each run, together with its
generation and CodeXGLUE evaluations, completes within a few GPU-hours on one
A100-80GB.

\begin{table}[h]
  \caption{Shared experimental configuration.}
  \label{tab:config}
  \centering
  \small
  \begin{tabular}{ll}
    \toprule
    Setting & Value \\
    \midrule
    Ablation & mean, over the prompt suite \\
    Necessity budget $k$ & 12 and 40 \\
    Minimality keep-fraction $\tau$ & 0.90 \\
    Specificity random sets $N$ & 30 \\
    Steering coefficients $\alpha$ & $\{0, 1, 2, 4, 8\}$ \\
    Aligned prompt pairs & 64 \\
    XLCoST prompt pairs & 128 \\
    Seed & 0 \\
    Hardware & $1\times$A100-80GB \\
    Precision & fp16 (bf16 for Qwen) \\
    \bottomrule
  \end{tabular}
\end{table}

\section{Untagged control}
\label{app:untagged}

The ``No tag'' column of Table~\ref{tab:preference} reports the mean $\dm$ over
ten bare instruction prompts (for example, ``Write a function that adds two
numbers.'') with no \texttt{\# language} tag. Negative means Python-leaning.
Twelve of fifteen settings stay negative and the three exceptions are all PHP,
which is consistent with PHP's markers overcoming a weak untagged prior in the
smaller models. The magnitudes are one to two orders below the tagged
separation, so the unconstrained default is real but faint. The Qwen family
shows a markedly stronger untagged default than CodeLlama or DeepSeek, and
within Qwen the 7B is stronger than the 14B, so unlike $S$ the untagged effect
does not grow monotonically with scale.

\section{Component depth and away-pointing shifts}
\label{app:depth}

\begin{figure}[h]
  \centering
  \includegraphics[width=0.62\linewidth]{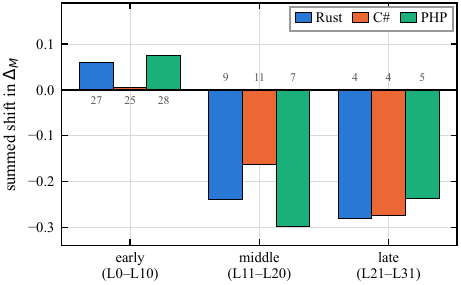}
  \caption{Summed individual patched shift of the $k=40$ candidate components in
  CodeLlama-7B, by depth band; numbers above or below each bar give how many
  components fall in that band. Early components are harmless, and the four to
  five late components carry as much negative shift as the whole middle band.
  The totals are nearly identical across the three languages, so depth explains
  the damage but not which setting ends up away-pointing (\S\ref{sec:wrongdir}).}
  \label{fig:depth}
\end{figure}

Figure~\ref{fig:depth} is the evidence behind \S\ref{sec:wrongdir}. It sums the
individual patched shift of every candidate component by depth band in
CodeLlama-7B, the model with the largest away-pointing result. Summing rather
than averaging is the right summary here because the combined patch transplants
all of them at once; note, though, that patching is not additive, so these sums
do not equal the combined shifts in Table~\ref{tab:intervention}.

\section{Extracted circuits per setting}
\label{app:circuits}

Table~\ref{tab:circuits} gives the circuit that greedy extraction returns in
each setting. A small circuit does not by itself imply specificity: several
single-component circuits in Table~\ref{tab:circuits} fail the test in
Table~\ref{tab:specificity}. The layers chosen share no pattern across models,
which is further evidence against a transferable syntax circuit.

\begin{table}[h]
  \caption{Circuit returned by greedy extraction (\S\ref{sec:setup}) per setting.
  ``Full candidate set'' means no compact subset reached 90\% of the full-set
  effect.}
  \label{tab:circuits}
  \centering
  \small
  \setlength{\tabcolsep}{4pt}
  \resizebox{\linewidth}{!}{\begin{tabular}{llrlrl}
\toprule
\multirow{2}{*}{Model} & \multirow{2}{*}{Target} & \multicolumn{2}{c}{Aligned, $k$=40} & \multicolumn{2}{c}{Instruct, $k$=12} \\
\cmidrule(lr){3-4}\cmidrule(lr){5-6}
 & & size & components & size & components \\
\midrule
CodeLlama-7B & Rust & 1 & \texttt{L9MLP} & 12 & \emph{full candidate set} \\
 & C\# & 40 & \emph{full candidate set} & 12 & \emph{full candidate set} \\
 & PHP & 1 & \texttt{L9MLP} & 12 & \emph{full candidate set} \\
\addlinespace[2pt]
DeepSeek-Coder-1.3B & Rust & 40 & \emph{full candidate set} & 6 & \texttt{L6MLP}, \texttt{L8MLP}, \texttt{L19MLP}, \texttt{L17MLP}, \ldots \\
 & C\# & 39 & \texttt{L18MLP}, \texttt{L10H1}, \texttt{L13MLP}, \texttt{L21MLP}, \ldots & 6 & \texttt{L13MLP}, \texttt{L21MLP}, \texttt{L12MLP}, \texttt{L5H3}, \ldots \\
 & PHP & 1 & \texttt{L5MLP} & 1 & \texttt{L3MLP} \\
\addlinespace[2pt]
DeepSeek-Coder-6.7B & Rust & 40 & \emph{full candidate set} & 2 & \texttt{L29MLP}, \texttt{L3MLP} \\
 & C\# & 2 & \texttt{L5MLP}, \texttt{L29MLP} & 1 & \texttt{L11MLP} \\
 & PHP & 2 & \texttt{L25MLP}, \texttt{L29MLP} & 12 & \emph{full candidate set} \\
\addlinespace[2pt]
Qwen2.5-Coder-7B & Rust & 2 & \texttt{L3MLP}, \texttt{L19H27} & 12 & \emph{full candidate set} \\
 & C\# & 5 & \texttt{L27H10}, \texttt{L0H6}, \texttt{L27H13}, \texttt{L18MLP}, \ldots & 12 & \emph{full candidate set} \\
 & PHP & 4 & \texttt{L23MLP}, \texttt{L22MLP}, \texttt{L27H13}, \texttt{L24MLP} & 12 & \emph{full candidate set} \\
\addlinespace[2pt]
Qwen2.5-Coder-14B & Rust & 40 & \emph{full candidate set} & 12 & \emph{full candidate set} \\
 & C\# & 1 & \texttt{L12MLP} & 12 & \emph{full candidate set} \\
 & PHP & 11 & \texttt{L33MLP}, \texttt{L36MLP}, \texttt{L43MLP}, \texttt{L18MLP}, \ldots & 1 & \texttt{L23MLP} \\
\bottomrule
\end{tabular}
}
\end{table}

\section{Full steering sweep}
\label{app:sweep}

Table~\ref{tab:sweep} gives the $\dm$ value at every coefficient for the $k=40$
base-model grid plotted in Figure~\ref{fig:steering}. Bold entries are
coefficients at which the most likely next token is target syntax. Reading the
table alongside Appendices~\ref{app:safe} and \ref{app:contam} shows the
problem with treating a metric crossing as a switch: several crossings occur at
coefficients where generation has already degraded.

\begin{table}[h]
  \caption{$\dm$ on the Python prompt across the steering sweep, base models at
  $k=40$. Bold marks coefficients where the most likely next token is target
  syntax.}
  \label{tab:sweep}
  \centering
  \small
  \begin{tabular}{llrrrrr}
\toprule
Model & Target & $\alpha$=0 & $\alpha$=1 & $\alpha$=2 & $\alpha$=4 & $\alpha$=8 \\
\midrule
CodeLlama-7B & Rust & -0.693 & -0.639 & -0.784 & -0.757 & -0.000 \\
 & C\# & -0.693 & -0.931 & +0.351 & +0.101 & +0.163 \\
 & PHP & -0.693 & -0.780 & -0.847 & -0.824 & -0.004 \\
\addlinespace[2pt]
DeepSeek-Coder-1.3B & Rust & -0.730 & -0.361 & +0.011 & \textbf{+0.870} & \textbf{+0.835} \\
 & C\# & -0.729 & \textbf{+0.682} & +0.160 & +0.042 & +0.073 \\
 & PHP & -0.729 & -0.670 & -0.623 & -0.646 & -0.000 \\
\addlinespace[2pt]
DeepSeek-Coder-6.7B & Rust & -0.779 & -0.917 & +0.001 & -0.000 & +0.000 \\
 & C\# & -0.779 & -0.681 & -0.569 & -0.427 & \textbf{+0.856} \\
 & PHP & -0.779 & -0.727 & -0.568 & \textbf{+0.335} & \textbf{+0.963} \\
\addlinespace[2pt]
Qwen2.5-Coder-7B & Rust & -0.910 & -0.883 & -0.848 & -0.770 & -0.691 \\
 & C\# & -0.910 & -0.867 & -0.742 & -0.173 & -0.116 \\
 & PHP & -0.910 & \textbf{+0.320} & \textbf{+0.824} & \textbf{+0.439} & \textbf{+0.336} \\
\addlinespace[2pt]
Qwen2.5-Coder-14B & Rust & -0.934 & -0.945 & -0.906 & -0.000 & -0.000 \\
 & C\# & -0.934 & -0.902 & -0.902 & -0.949 & -0.762 \\
 & PHP & -0.934 & -0.863 & \textbf{+0.605} & \textbf{+0.820} & \textbf{+0.668} \\
\bottomrule
\end{tabular}

\end{table}

\section{XLCoST detail}
\label{app:xlcost}

Table~\ref{tab:xlcost} separates the two terms of $\dm$ on the XLCoST suite.
Both are near zero, which is why the separation collapses: at the scored
position in naturally occurring code the model is not choosing between a
Python and a target-language keyword, so almost no probability mass falls on
either vocabulary. The comparison column repeats the aligned-prompt separation
for the same model and language.

\begin{table}[h]
  \caption{XLCoST prompts: both terms of $\dm$, the resulting separation, and the
  aligned-prompt separation for the same setting.}
  \label{tab:xlcost}
  \centering
  \small
  \begin{tabular}{llrrrr}
\toprule
Model & Target & $\dm(x_t)$ & $\dm(x_b)$ & $S$ & $S$ (aligned) \\
\midrule
CodeLlama-7B & C\# & +0.0101 & -0.0017 & +0.0117 & 1.551 \\
 & PHP & +0.0007 & +0.0000 & +0.0007 & 1.597 \\
DeepSeek-Coder-1.3B & C\# & +0.0001 & +0.0060 & -0.0060 & 1.488 \\
 & PHP & +0.0004 & -0.0022 & +0.0025 & 1.538 \\
DeepSeek-Coder-6.7B & C\# & +0.0036 & -0.0075 & +0.0111 & 1.625 \\
 & PHP & +0.0001 & -0.0070 & +0.0071 & 1.623 \\
Qwen2.5-Coder-7B & C\# & +0.0021 & -0.0021 & +0.0042 & 1.758 \\
 & PHP & +0.0027 & +0.0004 & +0.0024 & 1.758 \\
Qwen2.5-Coder-14B & C\# & +0.0082 & -0.0003 & +0.0086 & 1.828 \\
 & PHP & +0.0003 & +0.0000 & +0.0003 & 1.801 \\
\bottomrule
\end{tabular}

\end{table}

\section{Safe steering window}
\label{app:safe}

We call a coefficient \emph{safe} when it classifies every target prompt as
target-language and no control prompt as target-language. Table~\ref{tab:safe}
reports the largest safe coefficient per setting for the two Qwen models. The
window is strongly model- and language-dependent, which is the practical
counterpart of the negative steering result: no single strength is safe
everywhere. Qwen2.5-Coder-7B tolerates $\alpha=8$ on Rust and $\alpha=4$ on PHP,
while its C\# window collapses to $\alpha=0$. At 14B the pattern shifts again:
Rust is best left unsteered, while C\# and PHP have a genuine but narrow window
at $\alpha=2$.

\begin{table}[h]
  \caption{Largest safe steering coefficient per setting. ``Control FP'' is the
  rate at which control prompts are wrongly classified as target-language.}
  \label{tab:safe}
  \centering
  \small
  \begin{tabular}{llrrr}
\toprule
Model & Target & Safe $\alpha$ & Target success & Control FP \\
\midrule
Qwen2.5-Coder-7B & Rust & 8.0 & 1.00 & 0.00 \\
Qwen2.5-Coder-7B & C\# & 0.0 & 1.00 & 0.00 \\
Qwen2.5-Coder-7B & PHP & 4.0 & 1.00 & 0.00 \\
Qwen2.5-Coder-14B & Rust & 0.0 & 1.00 & 0.00 \\
Qwen2.5-Coder-14B & C\# & 2.0 & 1.00 & 0.00 \\
Qwen2.5-Coder-14B & PHP & 2.0 & 1.00 & 0.00 \\
\bottomrule
\end{tabular}

\end{table}

\section{Control contamination}
\label{app:contam}

Table~\ref{tab:contam} sweeps $\alpha$ on Qwen2.5-Coder-14B and reports the
target-prompt success rate together with the rate at which control prompts are
classified as target-language. C\# exposes the failure most sharply: at
$\alpha=2$ steering is effective and clean, but at $\alpha\geq4$ it injects
126--128 C\# markers into prompts that should contain none. The apparent metric
win at high $\alpha$ is an indiscriminate push, not language control.

\begin{table}[h]
  \caption{Qwen2.5-Coder-14B control-contamination stress test.}
  \label{tab:contam}
  \centering
  \small
  \begin{tabular}{lrrrr}
\toprule
Target & $\alpha$ & Target success & Control FP & Target markers in controls \\
\midrule
Rust & 0 & 1.00 & 0.00 & 0 \\
Rust & 4 & 0.00 & 0.00 & 0 \\
C\# & 2 & 1.00 & 0.00 & 0 \\
C\# & 4 & 1.00 & 1.00 & 126 \\
C\# & 8 & 1.00 & 1.00 & 128 \\
PHP & 2 & 1.00 & 0.00 & 2 \\
PHP & 8 & 1.00 & 0.50 & 6 \\
\bottomrule
\end{tabular}

\end{table}

\section{Leave-one-out stress test}
\label{app:loo}

Starting from the CodeLlama-7B/Rust top-12 set at $\alpha=1$, we remove one
component at a time and re-score the explicit Rust and control instruction
suite. Every removal, including dropping the one specific circuit, leaves marker
accuracy and F1 at $1.000$, and the mean next-token $\dm$ stays within $\pm0.02$
of the full-set value (Table~\ref{tab:loo}). No single component is
behaviourally load-bearing, which is what a redundant, multi-cue encoding
predicts.

\begin{table}[h]
  \caption{CodeLlama-7B/Rust leave-one-out steering at $\alpha=1$
  (representative rows; all twelve removals give accuracy/F1 $=1.000$).}
  \label{tab:loo}
  \centering
  \small
  \begin{tabular}{lrrr}
\toprule
Steering set & \#comp. & Acc./F1 & Mean $\dm$ \\
\midrule
no steering & 0 & 1.000 & 0.280 \\
full top-12 & 12 & 1.000 & 0.313 \\
$-$L0MLP & 11 & 1.000 & 0.312 \\
$-$L1MLP & 11 & 1.000 & 0.313 \\
$-$L29MLP & 11 & 1.000 & 0.299 \\
$-$L9MLP & 11 & 1.000 & 0.306 \\
$-$L7MLP & 11 & 1.000 & 0.311 \\
$-$L5MLP & 11 & 1.000 & 0.311 \\
$-$L4MLP & 11 & 1.000 & 0.310 \\
\bottomrule
\end{tabular}

\end{table}

\section{CodeXGLUE under steering}
\label{app:xglue}

We cast two CodeXGLUE tasks \citep{lu2021codexglue}, big-clone-bench clone
detection and defect detection, as zero-shot yes/no next-token decisions: the
prompt states the task and ends in ``Answer:'', and we compare the summed
probability of affirmative label tokens against negative ones on class-balanced
samples, so $0.50$ is chance. This is a negative control. If the steering
direction encoded task competence rather than surface syntax, steering would
move these scores. It does not (Table~\ref{tab:xglue}). The only movement is at
$\alpha=8$ on defect detection, where the label bias shifts and macro-F1 rises to
$0.491$ --- a redistribution of a degenerate predictor, not an accuracy gain.

\begin{table}[h]
  \caption{Zero-shot CodeXGLUE \citep{lu2021codexglue} under steering. Accuracy
  stays at chance; the $\alpha=8$ defect shift is a label-bias artifact.}
  \label{tab:xglue}
  \centering
  \small
  \begin{tabular}{lrrrr}
\toprule
Task & $\alpha$ & Acc. & F1 & Macro F1 \\
\midrule
clone & 0 & 0.500 & 0.667 & 0.333 \\
clone & 1 & 0.500 & 0.667 & 0.333 \\
clone & 8 & 0.500 & 0.667 & 0.333 \\
defect & 0 & 0.500 & 0.667 & 0.333 \\
defect & 1 & 0.500 & 0.667 & 0.333 \\
defect & 8 & 0.531 & 0.348 & 0.491 \\
\bottomrule
\end{tabular}

\end{table}

\end{document}